\documentclass[letterpaper]{article} 
\usepackage{aaai2026}  
\usepackage{times}  
\usepackage{helvet}  
\usepackage{courier}  
\usepackage[hyphens]{url}  
\usepackage{graphicx} 
\usepackage{natbib}  
\usepackage{caption} 
\usepackage{algorithm}
\usepackage{algpseudocode}

\usepackage{newfloat}
\usepackage{listings}
\DeclareCaptionStyle{ruled}{labelfont=normalfont,labelsep=colon,strut=off} 
\floatstyle{ruled}
\newfloat{listing}{tb}{lst}{}
\floatname{listing}{Listing}
\usepackage{amsmath}
\usepackage{amssymb}
\usepackage{multirow}
\usepackage{makecell}
\usepackage{booktabs} 
\usepackage[table,xcdraw]{xcolor}

\title{SAPO: Self-Adaptive Process Optimization Makes Small Reasoners Stronger}
\author{
    Kaiyuan Chen,
    Guangmin Zheng,
    Jin Wang\thanks{Corresponding author},
    Xiaobing Zhou,
    Xuejie Zhang
}
\affiliations{

    School of Information Science and Engineering, Yunnan University, Kunming, China \\
    chenkaiyuan@stu.ynu.edu.cn, gmzheng@mail.ynu.edu.cn, \{wangjin, zhouxb, xjzhang\}@ynu.edu.cn
}

\usepackage{bibentry}

\begin{document}

\maketitle

\begin{abstract}
Existing self-evolution methods overlook the influence of fine-grained reasoning steps, which leads to the reasoner-verifier gap. The computational inefficiency of Monte Carlo (MC) process supervision further exacerbates the difficulty in mitigating the gap. Motivated by the Error-Related Negativity (ERN), which the reasoner can localize error following incorrect decisions, guiding rapid adjustments, we propose a Self-Adaptive Process Optimization (SAPO) method for self-improvement in Small Language Models (SLMs). SAPO adaptively and efficiently introduces process supervision signals by actively minimizing the reasoner-verifier gap rather than relying on inefficient MC estimations. Extensive experiments demonstrate that the proposed method outperforms most existing self-evolution methods on two challenging task types: mathematics and code. Additionally, to further investigate SAPO's impact on verifier performance, this work introduces two new benchmarks for process reward models in both mathematical and coding tasks.
\end{abstract}


\section{Introduction}

Recent advances \cite{achiam2023gpt, guo2025deepseek} in Large Language Models (LLMs) highlight their superiority in complex multi-step planning, particularly with Chain-of-Thought (CoT) \cite{wei2022chain, kojima2022large}. Despite outperforming SLMs ($\le2{\rm{B}}$), LLMs' high computational and storage costs are prohibitive. Thus, developing efficient but strong SLMs for mobile devices is promising \cite{DBLP:conf/acl/MagisterMAMS23, chen2024mathematical}.

Previous approaches \cite{pang2024iterative,jiao-etal-2024-learning,guan2025rstar} have applied the self-evolutionary concept to LLMs to enhance reasoning performance, offering greater flexibility and efficiency. However, due to the poor instruction-following ability and reasoning performance, the general reward evaluation methods used on LLMs \cite{zheng2023judging,chen2024self,liu2025inference} are unsuitable for SLMs. In contrast, current classification-based discriminative methods are an efficient reward allocation approach better suited for the self-improvement of SLMs, typically a system composed of a reasoner and a verifier \cite{DBLP:journals/corr/abs-2402-06457, chen2025learning}, as shown in Figure \ref{fig:fig2}.

\begin{figure}[!t]
  \centering
    \includegraphics[width=0.85\linewidth]{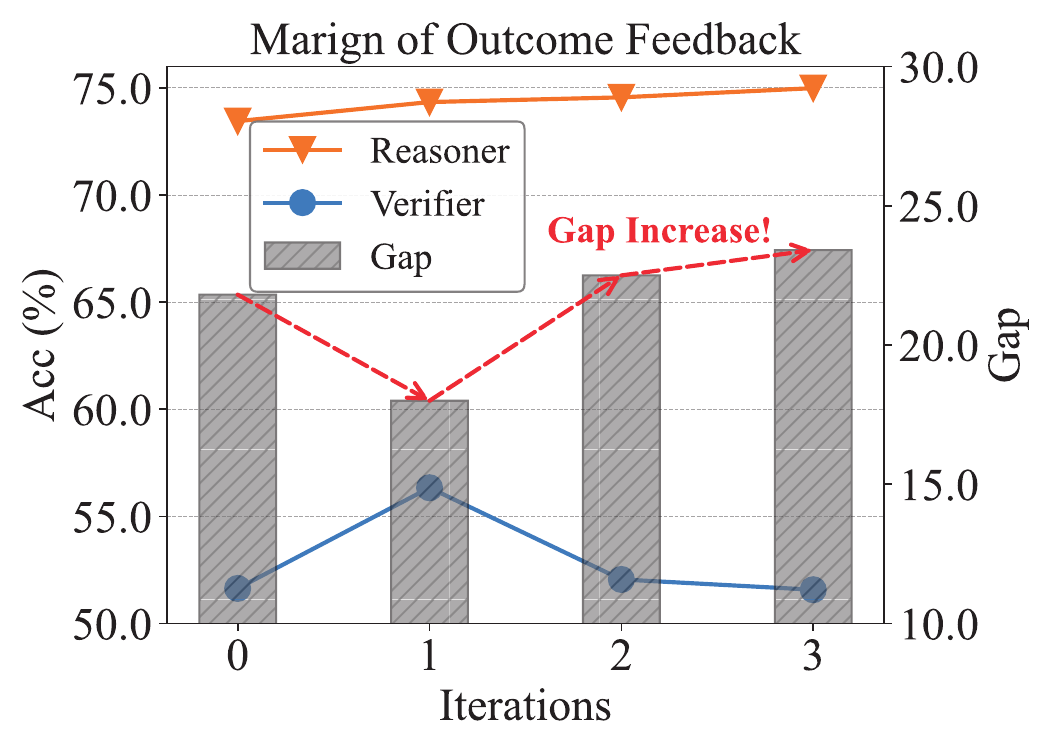}
    \label{fig:subfig1}
    \includegraphics[width=0.85\linewidth]{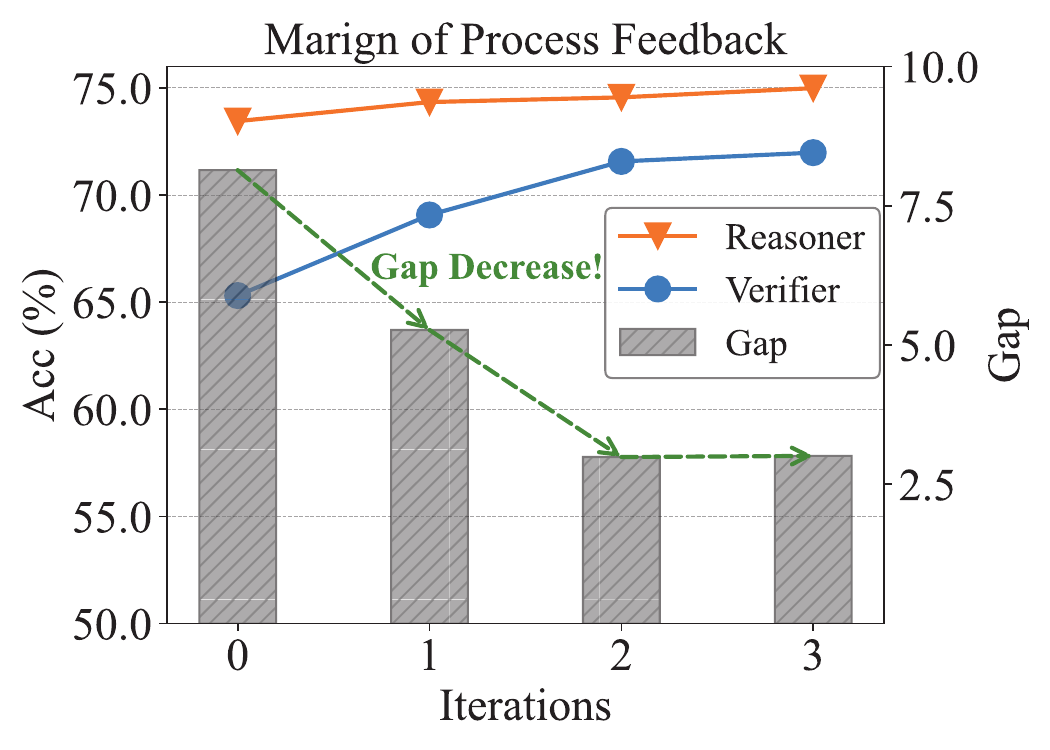} 
    \label{fig:subfig2}
  \caption{The performance and gap dynamics of the online reasoner and verifier across iterations in the step correction prediction of GSM\_Process. The selected baseline model is Qwen-2.5-0.5B, and the reasoner estimates step correctness using a MC approach.}
  \label{fig:fig1}
\end{figure}
\begin{figure*}[!t]
  \centering
  \includegraphics[width=0.8\linewidth]{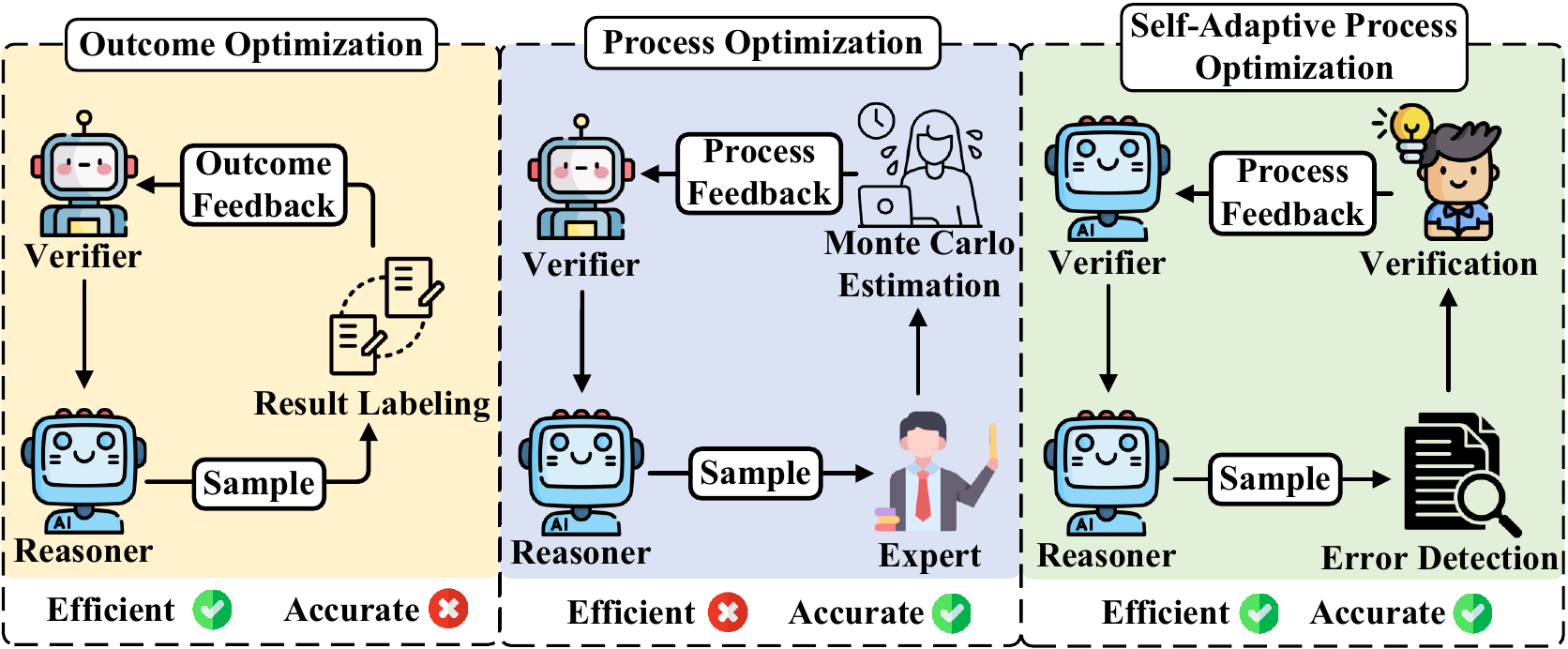}
  \caption{A comparison between the Self-Adaptive Process Optimization (SAPO) and the previous self-evolution framework. SAPO actively detects potential \textit{first error positions} to determine which steps need to be verified, rather than performing step-by-step rollout estimation.}
  \label{fig:fig2}
\end{figure*}

Nevertheless, most existing self-evolution methods still overlook feedback on fine-grained reasoning steps and instead opt for outcome rewards \cite{uesato2022solving, DBLP:conf/iclr/LightmanKBEBLLS24} as a more efficient form of supervision \cite{yu2025dapo,guo2025deepseek}, which inevitably leaves room for reward hacking \cite{liu2025understanding,yu2025dapo}. Thus, Monte Carlo-based process supervision transfers the posterior estimation of reasoning to the verifier model through step-level annotations, delivering superior performance \cite{jiao-etal-2024-learning, guan2025rstar, chen2025learning}.

To better understand the limitations of self-evolution methods guided by process-level feedback, this work investigates the bias issues arising within the \textit{exploration–exploitation} paradigm formed by the interaction between the reasoner and the verifier. Figure \ref{fig:fig1} (top) shows that, under multiple rounds of self-iteration, the absence of process supervision signals leads to an increasing reasoner-verifier gap. This gap undermines the verifier’s ability to assess the quality of reasoning paths accurately. In contrast, Figure \ref{fig:fig1} (bottom) shows that introducing online process supervision effectively reduces the reasoner-verifier gap.

However, the Monte Carlo-based process-supervised verifier \cite{DBLP:conf/acl/WangLSXDLCWS24,jiao-etal-2024-learning,guan2025rstar} is computationally expensive and highly inefficient. It poses a major bottleneck to scalability and practical application, as shown in Figure \ref{fig:fig2}. 
If each problem takes 8 steps and has 10 reasoning trajectories, labeling 10k problems requires 800,000 rollouts. 
Thus, achieving collaborative optimization between the inference engine and the verifier without significantly increasing supervisory costs is a key challenge in the current self-evolution methods.  

Therefore, we propose a novel Self-Adaptive Process Optimization (SAPO). This method enhances the efficiency of process supervision by introducing necessary step labels through localized error detection and correction. This new idea is motivated by the Error-Related Negativity (ERN) \cite{yeung2004neural}, whereby humans can spontaneously detect and localize errors shortly after making incorrect decisions, thereby guiding subsequent cognitive adjustments and behavioral corrections.

\begin{figure*}[!t]
  \centering
  \includegraphics[width=0.8\linewidth]{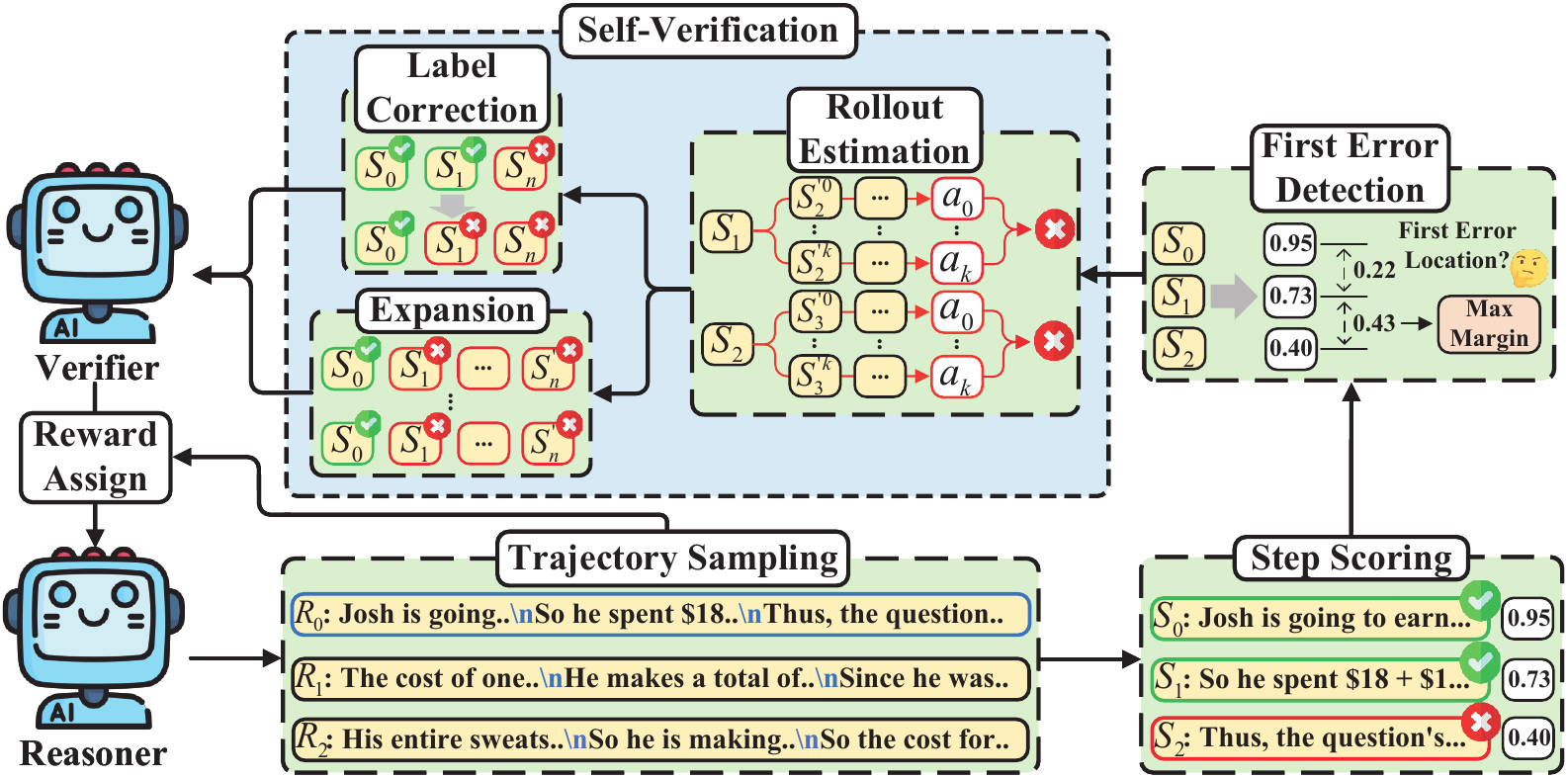}
  \caption{Overall framework diagram of Self-Adaptive Process Optimization (SAPO). The method adopts a self-iterative framework where the verifier pre-assigns step-level scores, error detection locates the first likely error, and the reasoner revisits it for posterior estimation. The corrected reasoning step labels then supervise the verifier, enabling the reasoner to self-optimize under more accurate process rewards.}
  \label{fig:fig3}
\end{figure*}

Refer to Figure \ref{fig:fig2} for a high-level understanding of SAPO. During the construction of process supervision, potential \textit{first error locations} \cite{uesato2022solving,DBLP:conf/iclr/LightmanKBEBLLS24} are identified through the changes and ordinality in reward scores. Then, the verifier is adaptively adjusted to minimize the discrepancies at these locations with the rollout estimates by the reasoner. The contributions of this work can be summarized as follows:
\begin{itemize}
\item [1)] 
This study proposes SAPO, a self-evolution paradigm that balances efficiency and performance by replacing inefficient step-by-step verification with first error detection and posterior estimation.
\item [2)]
To better evaluate the performance of different verifiers, this study introduces two datasets, GSM\_Process and MBPP\_Process, for reasoning step correctness verification in mathematics and code, respectively, based on first-error annotation. The Self-Adaptive PRM (SAPRM) also outperforms other reward models for process verification.
\item [3)]
Extensive experiments demonstrate that SAPO outperforms existing state-of-the-art self-evolution methods, e.g., GRPO \cite{shao2024deepseekmath}, on two challenging categories of multi-step reasoning tasks: mathematics and code. Moreover, SAPO demonstrates superior performance in addressing the bias between reasoning and verification.
\end{itemize}

\section{Preliminary}
\subsection{Problem Definition}
Given a question $q$, a reasoning model $M$ is to obtain its final answer $a$ through a certain reasoning trajectory $\tau$. Based on the Markov Decision Process (MDP) \cite{hao-etal-2023-reasoning,li2024process}, a reasoning trajectory can be defined as follows:
\begin{equation}
{s_{j + 1}} \sim M({s_{(0:j)}},q|\theta)
\end{equation}
\begin{equation}
\tau = \{ \langle s_0, c_0 \rangle, \langle s_1, c_1 \rangle, \ldots, \\
        \langle s_m, c_m \rangle \}
\end{equation}
where $\theta$ is the weight of the model $M$, ${s_i}$ is one of the immediate steps and ${s_{(0:j)}}$ represents an unfinished trajectory ending at ${s_j}$. Additionally, each step can be further assigned a corresponding correctness label $c \in \{ 0,1\}$.
\subsection{Process Reward Supervision}
An independent label can be assigned to each step, which can be manual. Recent works \cite{DBLP:conf/acl/WangLSXDLCWS24,jiao-etal-2024-learning} have adopted an MC estimation approach for automated labeling. Employing multiple rollouts starting from a certain step ${s_j}$ to estimate its correctness ${c_j}$, as follows:
\begin{equation}
\begin{split}
&\mathrm{Rollout}\bigl(M,q,s_{(0:j)}\bigr) = \bigl\{\,M\bigl(q,\,s_{(0:j)}^i \mid \theta\bigr)\bigr\}_{i=1}^K\\
&= \bigl\{\,\bigl(s_0^i,\dots,s_j^i,\;s_{j+1}^{i}\!'\,,\dots,s_m^{i}\!'\,,\;a_i\bigr)\bigr\}_{i=1}^K
\end{split}
\label{eq3}
\end{equation}
\begin{equation}
A = \{ {a_1},{a_2},...,{a_T}\}
\label{eq4}
\end{equation}
\begin{equation}
{c_j} = \left\{ {\begin{array}{*{20}{l}}
{1,}&{{\rm{if}}\;\exists\;{a_i} \in A,{a_i} = {a^*}}\\
{0,}&{{\rm{otherwise}}}
\end{array}} \right.
\label{eq5}
\end{equation}
where $K$ is the number of the sampling, and $m$ is the length of the completed trajectory. However, MC estimation is always computationally inefficient. Thus, a classification-based Process Reward Model (PRM) can be trained using Mean Square Error (MSE) loss, as follows:
\begin{equation}
{{\cal L}_{PRM}} = \frac{1}{n}\sum\limits_{i = 0}^n {\sum\limits_{j = 0}^{{\hat m_i}} {{{\left( {f(s_{(0:j)}^i;q) - {c^k}} \right)}^2}} }
\label{eq6}
\end{equation}
where $f$ denotes the predicted output of the classification head, $n$ denotes the total number of questions $q$, and ${\hat m_i}$ represents the total steps in the i-th reasoning trajectory. Typically, the PRM (or verifier $V$ ) can be initialized from $M$.

\section{Self-Adaptive Process Optimization}
\subsection{Reasoning Trajectory Sampling}
For a given question $q$, diverse reasoning trajectories can be obtained by the reasoner $M$ through high-temperature $T$ sampling \cite{yuan2023scaling}, as follows:
\begin{equation}
\mathrm{Sample}(M, q)
= \bigl\{\,\tau_i \mid \tau_i \sim M(q, T, \theta)\bigr\}_{i=1}^K
\end{equation}
$K$ denotes the number of sampled trajectories for a given problem. All sampled trajectories are deduplicated to ensure diversity, provided the minimum sampling count is met.

\subsection{Self-Adaptive Process Supervision}
\label{sec3.2}
\textbf{First Error Detection.} According to Equation (\ref{eq6}), the verifier $V$ can assign a reward score to any step ${s_j}$ within a sampled trajectory $\tau $, as follows:
\begin{equation}
{\hat c_j} = f({s_{(0:j)}};q) \in [0,1]
\end{equation}
according to previous works \cite{uesato2022solving, DBLP:conf/iclr/LightmanKBEBLLS24}, the first error position strategy is sufficient to provide effective process supervision signals. As shown in Figure \ref{fig:fig3}, the step with the maximum score difference $\Delta _j$ can be defined as the potential \textit{first error position}:
\begin{equation}
{\Delta _j} = {\hat c_{j-1}} - {\hat c_{j}}
\end{equation}
\begin{equation}
\hat t = argmax _{j \in \{ 1,2,...,m\} }{\Delta _j}
\end{equation}
where $\hat t$ denotes the potential first error position predicted by the $V$. Thus, step-level labels for a reasoning trajectory can also be pre-assigned:
\begin{equation}
\begin{split}
\tau  = 
&\{  \langle s_0^w,c_0^w \rangle ,..., \langle s_{\hat t-1}^w,c_{\hat t-1}^w \rangle ,\\
&\langle s_{\hat t}^l,c_{\hat t}^l \rangle ,..., \langle s_m^l,c_m^l \rangle \} 
\end{split}
\end{equation}
here, ${c^w} = 1$ and ${c^l} = 0$ indicate that the current step is correct and incorrect, respectively.

\noindent\textbf{Self-Verification.} The \textit{first error position} predicted by the verifier is not unbiased, as shown in Figure \ref{fig:fig3}. There may be a discrepancy between the predicted position  $\hat t$ and the gold position $t$ :(a)  If ${c_{\hat t-1}} = 1$ and ${c_{\hat t}} = 0$ : $\hat t = t$; (b) If ${c_{\hat t-1}} = 1$ and ${c_{\hat t}} = 1$ : $\hat t < t$; (c) If ${c_{\hat t-1}} = 0$ and ${c_{\hat t}} = 0$ : $\hat t > t$.

According to Equations (\ref{eq3}), (\ref{eq4}), and (\ref{eq5}), each case requires only two rollouts starting from $\hat t-1$ and $\hat t$ to verify,  which is significantly fewer than the step-by-step rollout estimation. Case (a) refers to the prediction being correct. For cases (b) and (c), the pre-assigned labels will be corrected as follows:
\begin{equation}
  \makebox[0pt][l]{(b)\quad}%
  \begin{split}
    \tau  =
    &\{\,\langle s_0^w,c_0^w\rangle,\dots,\langle s_{\hat t-1}^w,c_{\hat t-1}^w\rangle,\\
    &\langle s_{\hat t}^w,c_{\hat t}^w\rangle,\dots,\langle s_m^l,c_m^l\rangle\}
  \end{split}
\end{equation}
\begin{equation}
  \makebox[0pt][l]{(c)\quad}%
  \begin{split}
    \tau  =
    &\{  \langle s_0^w,c_0^w \rangle ,\dots, \langle s_{\hat t-1}^l,c_{\hat t-1}^l \rangle ,\\
    &\langle s_{\hat t}^l,c_{\hat t}^l \rangle ,\dots, \langle s_m^l,c_m^l \rangle \}
  \end{split}
\end{equation}

\noindent\textbf{Expansion.} Naively correcting a single-step error to reduce bias is suboptimal, as the corrected result may not generalize well to unseen cases. Fortunately, rollout inherently simulates diverse scenarios through sampling. Therefore, based on the \textit{first error location}, the extended trajectories can be retained to enhance the generalization of verification:
\begin{itemize}
\item[(i)]If ${c_{\hat t}} = 1$, it indicates that all extended trajectories with ${s_{(0:\hat t)}}$ as a prefix and a correct final result are correct at every step.
\item[(ii)]If ${c_{\hat t}} = 0$, it indicates that any suffix ${s_{(\hat t:m)}}$ starting from ${s_{\hat t}}$ leads to an incorrect final result being incorrect at every step.
\end{itemize}

\begin{table*}
    \centering
\begin{tabular}{ccccccc}
\toprule
\rowcolor[HTML]{EFEFEF} 
Model                      & \multicolumn{2}{c}{\cellcolor[HTML]{EFEFEF}Qwen-2.5-0.5B} & \multicolumn{2}{c}{\cellcolor[HTML]{EFEFEF}Llama-3.2-1B} & \multicolumn{2}{c}{\cellcolor[HTML]{EFEFEF}Gemma-2-2B} \\ \hline
\rowcolor[HTML]{EFEFEF} 
Method\textbackslash{}Task & GSM8K                       & MATH(OOD)                   & GSM8K                       & MATH(OOD)                  & GSM8K                      & MATH(OOD)                 \\ \hline
CoT                        & 28.51                       & 25.97                       & 5.31                        & 3.65                       & 19.86                      & 16.22                     \\
SFT                        & 34.19                       & 24.84                       & 22.14                       & 2.60                       & 39.12                      & 20.95                     \\
RFT                        & 37.83                       & 27.35                       & 26.31                       & {\underline{5.56}}                 & 45.34                      & 22.14                     \\
RFT+DPO                    & 8.26                        & 12.45                       & 23.88                       & 4.06                       & 41.02                      & 15.38                     \\
Online-RFT                 & 40.79                       & 28.85                       & 29.03                       & 5.45                       & {\underline{48.67}}                & \textbf{24.60}            \\
RPO                        & 41.55                       & 22.68                       & 29.26                       & 4.85                       & 45.41                      & 11.96                     \\ 
SFT+GRPO                   & \textbf{46.24}              & \textbf{34.53}              & 26.46                       & \textbf{5.92}              & 44.65                      & 24.36                     \\ \hline
SAPO-iter1                 & 36.99                       & 30.04                       & 29.41                       & 5.51                       & 46.10                      & 24.00                     \\
SAPO-iter2                 & 38.14                       & 31.48                       & {\underline{32.60}}                 & \textbf{5.92}              & 48.07                      & {\underline{24.36}}               \\
SAPO-iter3                 & {\underline{41.62}}                 & {\underline{31.72}}                 & \textbf{34.19}              & 5.45                       & \textbf{49.73}             & 24.24                     \\ \hline
\rowcolor[HTML]{EFEFEF} 
Method\textbackslash{}Task & MBPP                        & HumanEval(OOD)              & MBPP                        & HumanEval(OOD)             & MBPP                       & HumanEval(OOD)                 \\ \hline
CoT                        & 24.44                       & 25.60                       & 21.94                       & {\underline{19.51}}                & 22.61                      & 17.07                     \\
SFT                        & 29.32                       & 25.60                       & 24.01                       & {\underline{19.51}}                & 29.62                      & 21.34                     \\
RFT                        & 30.36                       & 26.82                       & 26.35                       & \textbf{21.34}             & 29.05                      & 18.90                     \\
RFT+DPO                    & 16.29                       & 20.73                       & 25.95                       & 17.68                      & 18.03                      & 16.46                     \\
Online-RFT                   & 34.00                       & 28.04                       & 28.25                       & 15.85                      & 34.20                      &20.73               \\
RPO                        & {\underline{36.63}}                 & \textbf{32.31}              & 28.45                       & 15.24                      & 34.93                      & 18.29               \\ 
SFT+GRPO                       & 35.20                       & 29.26                       & 25.55                       & 13.41                      & 33.40                      & \textbf{24.39}            \\ \hline
SAPO-iter1                 & 33.83                       & {\underline{31.09}}                 & 25.35                       & 18.90                      & 32.76                      & 22.56                     \\
SAPO-iter2                 & 34.06                       & 30.48                       & {\underline{28.82}}                 & \textbf{21.34}             & {\underline{35.17}}                & {\underline{23.78}}               \\
SAPO-iter3                 & \textbf{36.67}              & 30.48                       & \textbf{28.92}              & \textbf{21.34}             & \textbf{35.43}             & \textbf{24.39}            \\ \bottomrule
\end{tabular}
\caption{Comparative experimental results on multi-step reasoning tasks in both mathematics reasoning and code generation. Bold indicates the best performance, while underline represents the second best.}
\label{tab:table1}
\end{table*}

\subsection{Reasoning Optimization}
\label{sec3.3}
The verifier trained with self-adaptive process supervision can assign overall reward scores to sampled reasoning trajectories and construct a preference dataset, as follows:
\begin{equation}
r(\tau ) = \frac{{\sum\nolimits_{j = 0}^m {{{\hat c}_{j}}} }}{m} = \frac{{\sum\nolimits_{j = 0}^m {f({\hat s_{(0:j)}};q)} }}{m}
\end{equation}
\begin{equation}
{D_{pref}} = \{ {q_i},\tau _i^w,\tau _i^l\left|{r(\tau _i^w) - r(\tau _i^l) \ge \eta } \right.\}_{i=1}^n
\end{equation}
where $\eta $ denotes the threshold, while ${\tau ^w}$ and ${\tau ^l}$ represent the positive and negative samples, respectively. Given the preference data, SAPO employs the ORPO algorithm \cite{hong2024orpo} to achieve self-alignment of the reasoner $M$, as follows:
\begin{equation}
\begin{split}
{{\cal L}_{ORPO}} &=  \mathbb{E}{_{(q,{\tau ^w},{\tau ^l})}}\Bigl[{{\cal L}_{SFT}}(q,{\tau ^w})\\
 &- \beta \left( {\log \sigma \left( {\frac{{{\rm{odds}}({\tau ^w}\mid q)}}{{{\rm{odds}}({\tau ^l}\mid q)}}} \right)} \right)\Bigr]
\end{split}
\end{equation}
\begin{equation}
{\rm{odds(}}\tau \left| q \right.{\rm{)}} = \frac{{{P}(\tau \left| q \right.)}}{{1 - {P}(\tau \left| q \right.)}}
\end{equation}

\subsection{Iterative Self-Optimization}
The proposed SAPO follows an iterative \textit{exploration–exploitation} paradigm, where the reasoner is guided toward self-optimization by the verifier's progressively refined process supervision signals. The framework can be illustrated in Algorithm \ref{alg1}.

\begin{algorithm}[!t]
  \caption{Procedure of SAPO}
  \label{alg1}
  \begin{algorithmic}[1]
    \State Initialize: A pretrained SLM $M$; Original dataset $D_{\mathrm{org}}=\{(q_i,\tau_i)\}_{i=1}^N$
    \State $M_{0}\gets \mathrm{SFT}(M,\,D_{\mathrm{org}})$
    \State $D_{0}^{\mathrm{sample}}\gets \mathrm{Sample}(M_{0},\,D_{\mathrm{org}})$
    \State $D_{0}\gets D_{0}^{\mathrm{sample}}\cup D_{\mathrm{org}}$
    \State $D_{\mathrm{step\_label}}\gets \Omega\bigl(M_{0},\,\mathrm{Subset}(D_{0}^{\mathrm{sample}})\bigr)$
    \State $V_{0}\gets \mathrm{SFT}(M_{0},\,D_{\mathrm{step\_label}})$
    \For{$t = 1$ \textbf{to} $T$}
      \State $D_{t}^{\mathrm{sample}}\gets \mathrm{Sample}(M_{t-1},\,D_{t-1})$
      \State $D_{t}\gets D_{t}^{\mathrm{sample}}\cup D_{t-1}$
      \State $D_{\mathrm{score}}\gets \mathrm{Score}\bigl(V_{t-1},\,\mathrm{Subset}(D_{t}^{\mathrm{sample}})\bigr)$
      \State $D_{\mathrm{step\_detect}}\gets \mathrm{Detect}(D_{\mathrm{score}})$
      \State $D_{\mathrm{step\_label}}\gets \mathrm{Verify}(M_{t -1},\,D_{\mathrm{step\_detect}})$
      \State $V_{t}\gets \mathrm{SFT}(M_{t-1},\,D_{\mathrm{step\_label}})$
      \State $D_{\mathrm{perf}}\gets \mathrm{Score}(V_{t},\,D_{t})$
      \State $M_{t}\gets \mathrm{Align}(M,\,D_{\mathrm{perf}})$
    \EndFor
    \State \textbf{Return} The $M_{T}$ after iteration.
  \end{algorithmic}
\end{algorithm}

As indicated, \textit{Detect} and \textit{Verify} follow the strategies outlined before, and each verifier $V$ is initialized from the reasoner. Since the algorithm requires a trained verifier, we use binary estimation \cite{luo2024improve} (denoted as $\Omega$) to perform the initial step labeling.

\section{Experiments}
\subsection{Benchmark}
Two challenging multi-step reasoning benchmarks are selected to evaluate the model's reasoning capability: GSM8K \cite{cobbe2021training} for mathematical reasoning and MBPP \cite{austin2021program} for code generation. MATH \cite{hendrycks2021measuring} and HumanEval \cite{chen2021evaluating} are used to evaluate the Out-of-Domain (OOD) generalization on math and code tasks. Accuracy and Pass@1 \cite{chen2021evaluating} are evaluation metrics for math and code, respectively.

Previous studies used Best-of-N evaluation \cite{cobbe2021training} for verifier models, but it is unreliable for PRM \cite{zheng2024processbench,zhang2025lessons}. Instead, process-level verification provides a more dependable evaluation.

To enable step-wise verification in math and code tasks, we introduce two new benchmarks: GSM\_Process and MBPP\_Process. For each question in GSM8K and MBPP, different models generated two reasoning paths. GPT-4o \cite{achiam2023gpt} was used to annotate the first incorrect step in each path, with results filtered for reliability. These benchmarks contain 3,786 and 1,499 examples, and evaluate verifiers based on step correction prediction accuracy.

\subsection{Baselines}
Except for Chain-of-Thought (CoT) \cite{wei2022chain} and Supervised Fine-Tuning (SFT), several strong self-evolution methods are also implemented as baselines, including: Rejection Sampling Fine-Tuning (RFT) \cite{yuan2023scaling}, RFT combined with Direct Preference Optimization \cite{rafailov2023direct} (RFT+DPO), Reasoning Preference Optimization (RPO) \cite{pang2024iterative}, Group Relative Policy Optimization \cite{shao2024deepseekmath} after SFT (SFT+GRPO). Additionally, ORM \cite{DBLP:conf/iclr/LightmanKBEBLLS24}, OmegaPRM \cite{luo2024improve}, and ShepherdPRM \cite{DBLP:conf/acl/WangLSXDLCWS24} are implemented as baselines for verifiers' comparison.

\subsection{Setup}
Qwen-2.5-0.5B \cite{DBLP:journals/corr/abs-2412-15115}, Llama-3.2-1B \cite{grattafiori2024llama}, and Gemma-2-2B \cite{team2024gemma} serve as evaluation backbones. For SAPO, distinct reasoning trajectories are sampled per question per iteration, using the same model as the verifier. Full fine-tuning is applied to Qwen-2.5-0.5B and Llama-3.2-1B, while QLoRA \cite{DBLP:conf/nips/DettmersPHZ23} fine-tunes Gemma-2-2B for efficiency. The publicly available Unsloth framework is used to accelerate training and inference, and all experiments are run on two NVIDIA 3090 GPUs.

\begin{figure*}[!t]
  \centering
  \includegraphics[width=0.75\linewidth]{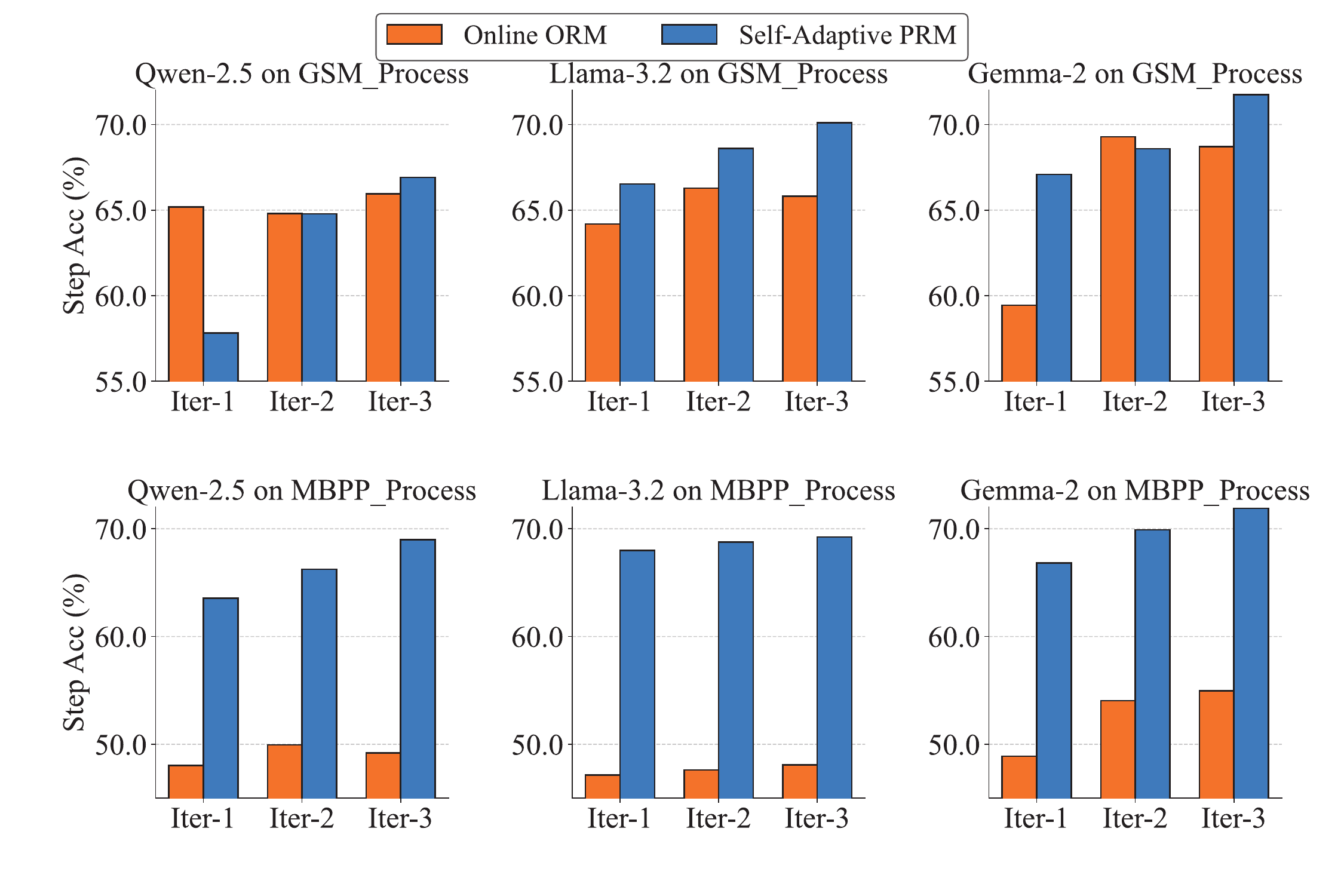}
  \caption{The performance evolution of SAPRM across different tasks (GSM\_Process and MBPP\_Process) and models over iterations. The online Outcome Reward Model (ORM) is used as a baseline for comparison.}
  \label{fig:fig4}
\end{figure*}

\begin{figure}[!t]
   \centering
  \includegraphics[width=0.9\columnwidth]{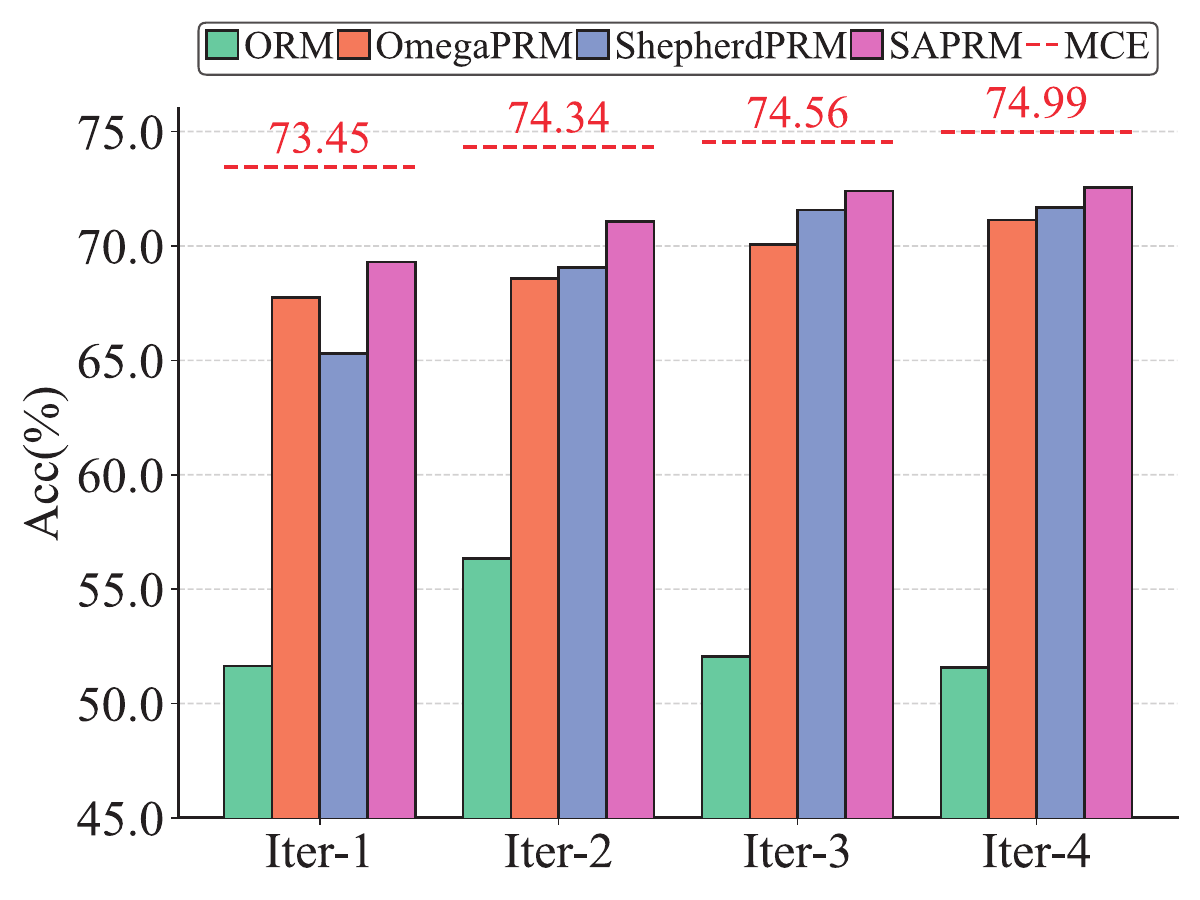}
  \caption{Multi-round iterative performance comparison of different reward models on GSM\_Process. Monte Carlo Estimation (MCE) serves as the upper bound of verification performance achievable by the reasoner, and Qwen-2.5-0.5B is used as the base model.}
  \label{fig:fig6}
\end{figure}

\subsection{Comparison of Different Self-Evolution Methods}
Table \ref{tab:table1} shows that SAPO consistently outperforms most other methods across different models and tasks, both in-domain and out-of-domain. Moreover, reasoning performance of SAPO can further improve with more iterations.

Applying DPO after RFT to align positive and negative sample results in worse performance, consistent with previous finding \cite{feng2024towards}. And effective improvement can be achieved by emphasizing positive examples during the iterative process, as demonstrated by RPO \cite{pang2024iterative} and ORPO \cite{hong2024orpo}.

Although Qwen-2.5-0.5B trained with GRPO achieves the best performance on math tasks, this improvement is highly model-dependent. When the base model has weaker capabilities (e.g., LLaMA or Gemma) or when applied to code tasks, GRPO does not outperform other methods. Moreover, without SFT, training SLMs with GRPO struggles to converge to high-reward.

\subsection{Comparison of Different Verifiers}
Figure \ref{fig:fig4} presents the performance trends of SAPRM on mathematical and code tasks. With only local step corrections introduced, SAPRM achieves progressively improved process verification performance across iterations, often outperforming ORM, particularly on code verification. In contrast, Online ORM exhibits a clear performance ceiling—its verification accuracy tends to decline after reaching a peak. Moreover, SAPRM performs better with larger models.

To further investigate the reasoning-verification bias under different approaches, we also implemented two commonly used baselines, OmegaPRM and ShepherdPRM, for comparison, as shown in Figure \ref{fig:fig6}. Additionally, we used step prediction accuracy based on MCE starting from each step of the reasoning process as an upper bound for each iteration. Figure \ref{fig:fig6} shows that SAPRM achieves lower bias on GSM\_Process than other verifier models, despite only verifying and correcting local information.

\section{Further Analysis}
\subsection{Analysis of Process Supervision Efficiency}
As described before, Self-Adaptive Process Supervision (SAPS) enhances efficiency by identifying and verifying only the first potential error, thereby avoiding unnecessary step-by-step rollout. To validate the effectiveness of the proposed method, we compare SAPS with two current process labeling approaches, including Shepherd and Omega, on processing supervision signal annotation efficiency. The comparison uses FLOPs and elapsed time (in seconds) on different tasks. The evaluation was conducted using Qwen-2.5-0.5B on a single NVIDIA RTX 3090 GPU.
For the code tasks, compilation time was also taken into account.

\begin{figure*}[!t]
  \centering
  \includegraphics[width=0.7\linewidth]{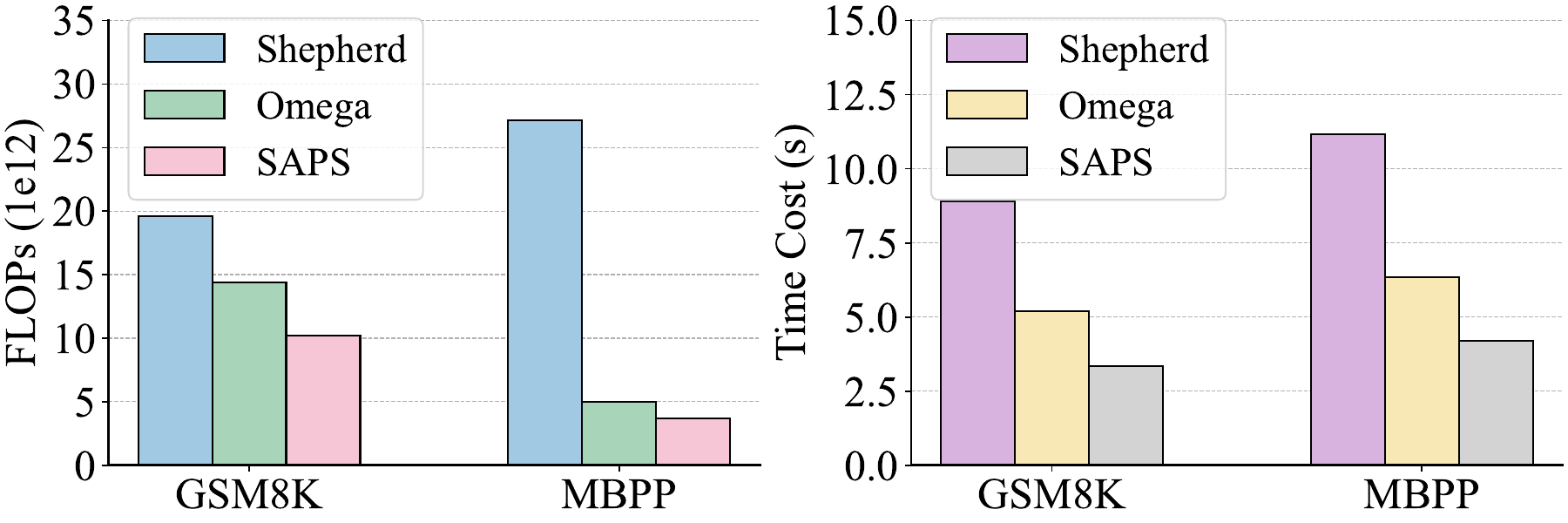}
  \caption{Comparison of efficiency in process labeling methods. The figure illustrates the average FLOPs and time (in seconds) required to label a single sample under different process supervision methods.}
  \label{fig:fig5}
\end{figure*}

Figure \ref{fig:fig5} shows that, compared to Shepherd, which relies on step-wise rollout for verification, SAP achieves a 2–3x improvement in FLOPs and time consumption. Moreover, SAPS outperforms Omega, which uses binary estimation to locate the predicted first error position.

\subsection{Analysis of Self-Verification Errors}
Self-verification error rate reflects the consistency between the reasoner and the verifier.
Additionally, the efficiency of SAPO also depends on the number of mismatches or errors that occur when the reasoner is used to verify the labels pre-assigned by the verifier.
Thus, it is worth exploring the relationship between the reasoner and the verifier across different iterations.
Specailly, Pre-assigned labels are considered correct in only three cases: (1) correct result with all correct steps, (2) incorrect result with all incorrect steps, or (3) incorrect result with correctly locating the first error.

To avoid interference from varying samples, we uniformly adopt the reasoning trajectories used by three different models from GSM\_Process and MBPP\_Process for self-verification error evaluation. The analysis is conducted to determine whether the reasoner and verifier are aligned.

Table \ref{tab:table2} shows that self-verification performs best when the reasoner and verifier are from the same round (e.g., \textit{V3-R3}), while mismatched pairs lead to higher error rates, highlighting the importance of online synchronization. Notably, although  \textit{V1-R1} is also synchronized, its higher error rate compared to  \textit{V3-R3} suggests that iteration helps close the gap between the verifier and reasoner.

\subsection{Ablation Study Analysis}
In the ablation study, \textit{w/o PF} removes process feedback from verifier updates, assigning step labels only by final result; \textit{w/o RM} excludes the verifier, using only result correctness to label samples; \textit{w/o DV} replaces verifier-guided correction with random step correction; \textit{w/o EP} disables verification expansion. Table \ref{tab:table3} shows that removing any component leads to varying degrees of performance degradation, demonstrating their necessity. Moreover, \textit{w/o PF} and \textit{w/o RM} have distinct impacts despite both relying solely on outcome feedback. \textit{w/o PF} leads to a more significant overall performance degradation due to its effect on the PRM.

\begin{table}[!t]
    \centering
  \setlength{\tabcolsep}{1mm}
    \begin{tabular}{cccc}
    \toprule
		Model &Qwen-2.5-0.5B &Llama-3.2-1B  &Gemma-2-2B    \\
    \midrule
    Task &\multicolumn{3}{c}{GSM8K} \\
    \midrule
    \textit{V1-R1}   &58.59  &60.34 &54.51 \\
    \textit{V1-R3}   &58.66  &61.49 &56.48 \\
    \textit{V3-R3}   &\textbf{54.63}  &\textbf{54.46} &\textbf{51.36} \\
    \midrule
    Task &\multicolumn{3}{c}{MBPP} \\
    \midrule
    \textit{V1-R1}   &67.48  &78.22 &67.48 \\
    \textit{V1-R3}   &68.85  &76.96 &67.31 \\
    \textit{V3-R3}   &\textbf{67.08}  &\textbf{76.21} &\textbf{63.29} \\
    \bottomrule
	\end{tabular}
  \caption{The self-verification error rates (\%) obtained by the different reasoner and verifier. \textit{V1-R3} indicates that the verifier from round 1 is used to pre-assign labels, while the reasoner from round 3 is used to perform self-verification.}
  \label{tab:table2}
\end{table}

\section{Related Work}
\noindent\textbf{Self-Evolution of Language Reasoning.} Reasoning models have long been a central focus of research \cite{guo2025deepseek, shen2025insight, shen2025flow, zheng2024enhancing, 11209812, yuan2023joint, yuan2025collaborative}. And self-evolution is central to recent reasoning methods. STaR \cite{zelikman2022star} and RFT \cite{yuan2023scaling} apply supervised learning to model-generated traces. RPO \cite{pang2024iterative} and V-STaR \cite{DBLP:journals/corr/abs-2402-06457} both use offline reinforcement (e.g., DPO) to align positive and negative samples over multiple rounds, while V-STaR further integrates a verifier to internalize this alignment. Current R1-like self-reflective reasoning \cite{liu2025understanding,yu2025dapo} that centers around algorithms like GRPO \cite{shao2024deepseekmath} or PPO \cite{schulman2017proximal} can also be categorized as self-evolution. Specifically, these methods iteratively adjust the model through online sampling to produce higher-reward outputs.

While self-evolution approaches have improved pretrained models' reasoning, they rely mainly on outcome feedback, which leaves room for reward hacking during alignment \cite{liu2025understanding}.

\begin{table}[!t]
	\centering
  \setlength{\tabcolsep}{1mm}
    \begin{tabular}{lcc}
		\toprule
		Method\textbackslash Model    &Qwen-2.5-0.5B   &Llama-3.2-1B \\
    \midrule
    SAPO    &\textbf{41.62}   &\textbf{34.19} \\
    \midrule
    \textit{w/o PF}   &39.65  &32.37 \\
    \textit{w/o DV}   &40.86  &31.69 \\
    \textit{w/o RM}   &40.71  &32.75 \\
    \textit{w/o EP}   &40.33  &33.73 \\
    \bottomrule
	\end{tabular}
  \caption{Ablation study on GSM8K. \textit{PF} denotes process feedback, \textit{DV} stands for error detection and self-verification, \textit{RM} represents the reward model (or verifier), and \textit{EP} refers to expansion during verification.}
  \label{tab:table3}
\end{table}

\noindent \textbf{Process-guided Long-Chain Reasoning.} Compared to outcome-level supervision, step-level process supervision has gained attention for offering denser feedback signals. Early studies \cite{uesato2022solving,DBLP:conf/iclr/LightmanKBEBLLS24} manually labeled the correctness of individual reasoning steps to train a PRM. Recent works \cite{DBLP:conf/acl/WangLSXDLCWS24,luo2024improve,jiao-etal-2024-learning} have focused on automating step verification using Monte Carlo-based rollout estimation.

Research based on process supervision generally falls into two categories: (1) training-free methods \cite{hao-etal-2023-reasoning,guan2025rstar} using Monte Carlo Tree Search (MCTS) for test-time performance scaling, and (2) self-evolutionary approaches \cite{jiao-etal-2024-learning,chen2025learning} that improve reasoning via offline PRMs in an \textit{exploration-exploitation} framework. However, these self-evolution methods still face inefficient training and weak alignment between rewards and reasoning steps.

\section{Conclusion}
This paper proposed the Self-Adaptive Process Optimization (SAPO) method to improve multi-step reasoning performance in SLMs. Specifically, the method gradually enhances reasoning performance by adapting to the bias between the reasoner and the verifier. Extensive experimental results show that the proposed method outperforms existing self-evolution methods and is more efficient in step labeling than previous approaches.

\section{Acknowledgments}
This work was supported by the National Natural Science Foundation of China (NSFC) under Grant Nos. 61966038 and 62266051, and the Postgraduate Practice and Innovation Foundation of Yunnan University under Grant No. ZC-242410094.
We would like to thank the anonymous reviewers for their constructive comments.

\bibliography{aaai2026}

\newpage
\section{A. Experimental Details}
\subsection{A.1. Benchmark}
\label{apd:A.1}
The evaluation of reasoning models primarily focuses on two types of tasks: mathematical reasoning and code generation, as detailed below:

\textbf{GSM8K} \cite{cobbe2021training} is a dataset consisting of grade school math word problems, each requiring 2 to 8 steps of reasoning to solve. The publicly available dataset contains 7,473 training samples and 1,319 test samples. During the experiments, this benchmark was used to evaluate the in-domain performance of mathematical reasoning.

\textbf{MATH} \cite{hendrycks2021measuring} is a highly challenging dataset consisting of mathematical competition problems that span various domains such as calculus, geometry, and algebra. Solving each problem typically requires specialized and advanced mathematical knowledge.
Since the standard answers in this dataset are often not in a simple floating-point numerical format, to avoid performance bias caused by inconsistencies in final answer formats when evaluating models trained on GSM8K, we filtered the test set and retained only those problems that could be answered with a floating-point number. As a result, 1,691 problems from MATH were selected for out-of-domain evaluation in mathematical reasoning.

\textbf{MBPP} \cite{austin2021program} is a benchmark consisting of approximately 1,000 Python programming problems, with 374 samples in the training set and 500 in the test set. Each problem in the dataset includes three test cases, and a code solution is 
considered correct only if it passes all of them. This task is used for in-domain evaluation of code generation.

\textbf{HumanEval} \cite{chen2021evaluating} is a code evaluation benchmark consisting of only 164 programming problems, each with corresponding test cases. In our experiments, the problem formats in this dataset were adjusted to ensure that models trained on MBPP are not affected by differences in input formatting.

\begin{table*}[!t]
  \centering
  \small
  \setlength{\tabcolsep}{4pt}
  \resizebox{\textwidth}{!}{%
    \begin{tabular}{l
                    cccc|
                    cccc|
                    cccc}
      \toprule
         Model 
         & \multicolumn{4}{c}{Qwen-2.5-0.5B} 
         & \multicolumn{4}{c}{Llama-3.2-1B} 
         & \multicolumn{4}{c}{Gemma-2-2B} \\
      \cmidrule(lr){2-13}
        Method
        & \multicolumn{2}{c}{SFT} 
        & \multicolumn{2}{c}{ORPO} 
        & \multicolumn{2}{c}{SFT} 
        & \multicolumn{2}{c}{ORPO} 
        & \multicolumn{2}{c}{SFT} 
        & \multicolumn{2}{c}{ORPO} \\
      \cmidrule(lr){2-5}
      \cmidrule(lr){6-9}
      \cmidrule(lr){10-13}
      Task
      & GSM8K & MBPP & GSM8K & MBPP
      & GSM8K & MBPP & GSM8K & MBPP
      & GSM8K & MBPP & GSM8K & MBPP \\
      \midrule
      Learning\_Rate 
        & 5e-5  & 2e-5  & 8e-6  & 5e-6  
        & 5e-5  & 3e-5  & 5e-5  & 2e-5  
        & 1e-4  & 5e-5  & 8e-5  & 2e-5  \\
      Optimizer      & \multicolumn{4}{c|}{AdamW} & \multicolumn{4}{c|}{AdamW} & \multicolumn{4}{c}{AdamW} \\
      Warmup\_Ratio  & \multicolumn{4}{c|}{0.1} & \multicolumn{4}{c|}{0.1} & \multicolumn{4}{c}{0.1} \\
      Batch\_Size    & 24    & 20    & 32    & 32    
                     & 24    & 20    & 32    & 32    
                     & 24    & 20    & 32    & 32    \\
      Dropout\_Ratio & \multicolumn{4}{c|}{0.01}     
                     & \multicolumn{4}{c|}{0.01}     
                     & \multicolumn{4}{c}{--}     \\
      OR\_Weight     & \multicolumn{4}{c|}{0.1}     
                     & \multicolumn{4}{c|}{0.1}     
                     & \multicolumn{4}{c}{0.1}   \\
      Lora\_Alpha    & \multicolumn{4}{c|}{--} & \multicolumn{4}{c|}{--} & \multicolumn{4}{c}{128} \\
      Lora\_R        & \multicolumn{4}{c|}{--} & \multicolumn{4}{c|}{--} & \multicolumn{4}{c}{256} \\
      Lora\_Dropout  & \multicolumn{4}{c|}{--} & \multicolumn{4}{c|}{--} & \multicolumn{4}{c}{0.05} \\
      \bottomrule
    \end{tabular}%
  }
  \caption{Hyperparameter settings for different tasks and models. Among them, the optimizer, the weight of the loss for aligning positive and negative examples in ORPO (OR\_Weight), and the warm-up ratio are kept consistent across all configurations.}
  \label{tab:table4}
\end{table*}

\subsection{A.2. Detail of Setup}
\label{apd:A.2}
The experiments are conducted on three commonly used SLMs: Qwen-2.5-0.5B \cite{DBLP:journals/corr/abs-2412-15115}, Llama-3.2-1B \cite{grattafiori2024llama}, and Gemma-2-2B \cite{team2024gemma}. Furthermore, the experimental setup can be organized according to the overall framework of SAPO, as follows:

\noindent\textbf{Initialization.} The iterative algorithm of SAPO requires two key components: a reasoner capable of performing multi-step CoT reasoning for a given problem, and a verifier that can predict step-level scores based on the given problem and the corresponding CoT. 

Therefore, for the initialization of the reasoner, we conduct supervised fine-tuning on the original GSM8K and MBPP training datasets, respectively. For the initialization of the verifier, since its training requires step-level labels, we adopt Omega's binary search-based step annotation method \cite{luo2024improve} to initialize the step labels for the first-round sampled trajectories. In practice, there are various alternatives for the initializing step labels—for example, labeling step correctness based on the final result. However, we did not explore additional initialization strategies in this work and leave such investigations for the future. The more hyperparameters for initialization can be found in Table \ref{tab:table4}.

\begin{table}[!t]
    \centering
  \setlength{\tabcolsep}{1mm}
    \resizebox{\linewidth}{!}{
    {\fontsize{16}{20}\selectfont
    \begin{tabular}{ccccccc}
    \toprule
     Model &\multicolumn{2}{c}{Qwen-2.5-0.5B} &\multicolumn{2}{c}{Llama-3.2-1B}  &\multicolumn{2}{c}{Gemma-2-2B}    \\
    \midrule
    Task &GSM8K &MBPP &GSM8K &MBPP &GSM8K &MBPP\\
    \midrule
    Iter-0   &7.4  &2.2 &7.4  &2.2 &7.4  &2.2 \\
    Iter-1   &25.6  &10.0 &30.2 &10.0  &28.8 &10.6 \\
    Iter-2   &44.9  &15.1 &50.4 &21.9  &45.2 &21.7 \\
    Iter-3   &58.3  &33.8 &74.6 &41.8  &75.0 &32.6 \\
    \bottomrule
	\end{tabular}
  }}
  \caption{The number (k) of positive and negative samples obtained in each iteration under SAPO across different models and tasks.}
  \label{tab:table5}
\end{table}

\noindent\textbf{Sampling.} In each iteration, the newly updated model collects diverse reasoning trajectories through high-temperature sampling. Here, "diverse" refers to samples that are deduplicated by comparing them with those collected in previous iterations. Specifically, for math and coding tasks, deduplication is performed based on the order of arithmetic operations and edit distance, respectively. By default, the sampling temperature is set to 1 for mathematical reasoning and 0.7 for coding generation.

\noindent\textbf{Step Scoring and First Error Detection.} The implementation follows the method described in the framework, with the default threshold for determining whether a reasoning step is correct or incorrect set to 0.5.

\noindent\textbf{Self-Verification.} To balance efficiency and diversity, the experiment does not perform verification on trajectories sampled from all GSM8K problems. Instead, K-means clustering ($K=8$) is applied to partition the problems, and the 100 problems closest to each cluster center are selected as the fixed set for verification in each iteration. However, for MBPP, the verification is conducted on all 374 problems in the dataset.

\noindent\textbf{Optimization.} As described before, ORPO is applied to achieve self-alignment over reasoning trajectories of high and low reward values. This choice of algorithm is motivated by prior observations that combining SFT on positive samples with alignment over both positive and negative samples is better suited for reasoning tasks. The ORPO hyperparameter settings for different models can be found in Table \ref{tab:table4}. In addition, the amount of data used in each iteration is summarized in Table \ref{tab:table5}.

\subsection{A.3. Implementation of the Baselines}
\label{apd:A.3}
\noindent\textbf{CoT.} For CoT, the 8-shot prompt template from previous work \cite{wei2022chain} is adopted to guide the model in performing long-chain reasoning across all experiments.

\noindent\textbf{SFT.} Providing the model with question-answer pairs \cite{cobbe2021training,DBLP:conf/acl/MagisterMAMS23} can also effectively guide CoT reasoning. In our experiments, we conduct SFT for mathematical reasoning on GSM8K and code generation on MBPP using their original training datasets.

\noindent\textbf{RFT.} The model is first initialized via supervised fine-tuning (SFT) on the original dataset. It then collects diverse reasoning trajectories through high-temperature sampling ($T=1$), followed by filtering based on answer correctness to construct supervision data for RFT \cite{yuan2023scaling}. To ensure a fair comparison, at least 24 distinct reasoning trajectories are collected for each question.

\noindent\textbf{RFT+DPO.} Since RFT only leverages self-sampled positive examples, we further categorized the sampled trajectories based on their final correctness and applied DPO \cite{rafailov2023direct} on top of the RFT model to achieve reasoning alignment.

\noindent\textbf{Online-RFT.} RFT can be further implemented in an online manner \cite{DBLP:journals/corr/abs-2402-06457}, eliminating the need to collect a large amount of data in the initial iteration. Specifically, the model fine-tuned via RFT in the previous round samples 8 diverse reasoning trajectories per question with a temperature of 1 to construct a training dataset for the next iteration. In the experiments, the number of iterations for Online-RFT is set to 3, which is consistent with the configuration used in SAPO.

\noindent\textbf{RPO.} To address the issue in DPO alignment where the generation probabilities of both positive and negative samples may decrease, RPO incorporates both the SFT loss on positive samples and the alignment loss on positive-negative pairs. Similarly, RPO \cite{pang2024iterative} is an iterative algorithm. In our experiments, the number of iterations for RPO is also set to 3, and in each round, 8 distinct reasoning trajectories are collected per problem.

\noindent\textbf{GRPO.} Through online sampling and a training paradigm that maximizes group advantages, GRPO \cite{guo2025deepseek,shao2024deepseekmath} has recently been regarded as an effective approach to stimulate self-reflection in LLMs and enhance their long-chain reasoning capabilities.

\begin{figure}[!t]
  \includegraphics[width=\columnwidth]{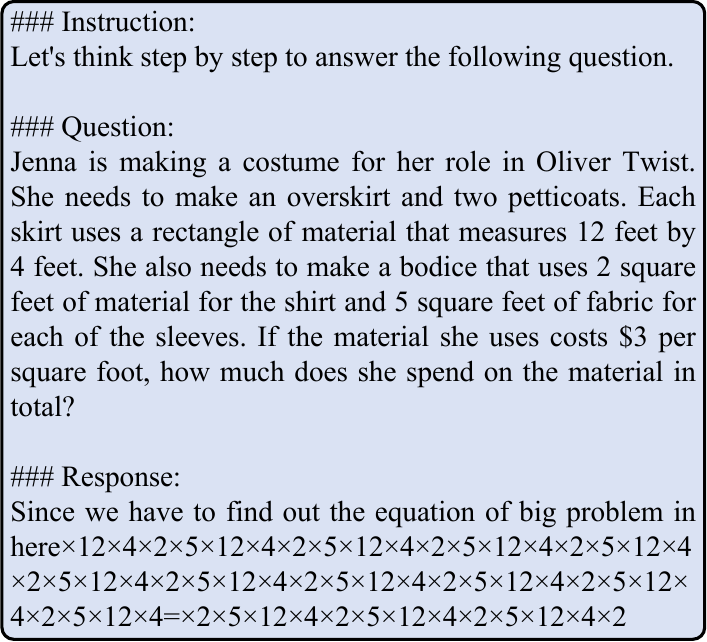}
  \caption{The sampled reasoning response from Llama-3.2-1B on GSM8K after late-stage GRPO training exhibits clear repetition and errors.}
  \label{fig:fig10}
\end{figure}

However, in our experiments, we found that directly applying GRPO to SLMs with weak instruction-following capabilities (such as Qwen-2.5-0.5B) led to noisy and unstable outputs that failed to converge, as illustrated in Figure \ref{fig:fig10}. The learning rate for GRPO was uniformly set to 5e-6 with a batch size of 64, and training was conducted over 5 iterations using 8 sampled trajectories per problem. In addition, only the result correctness (or whether the test cases are passed) is used as the reward function. Due to the substantial computational overhead of the GRPO algorithm itself, QLoRA is adopted (with rank=64, alpha=32) in combination with the Unsloth framework for efficient fine-tuning.

\section{B. Evaluation of Verifier}
\label{apd:B}
Although existing works \cite{cobbe2021training,DBLP:conf/iclr/LightmanKBEBLLS24,li2024process,yuan2024free} often use Pass@k or Accuracy under the BoN sampling setting to evaluate the reliability of verification models, assessing PRM performance solely based on outcomes inevitably introduces significant bias \cite{zhang2025lessons,zheng2024processbench}. This is primarily because the reliability of intermediate reasoning steps cannot be accurately inferred from the correctness of the final answer \cite{zhang2025lessons}.

\begin{figure*}[!t]
  \centering
  \includegraphics[width=0.85\linewidth]{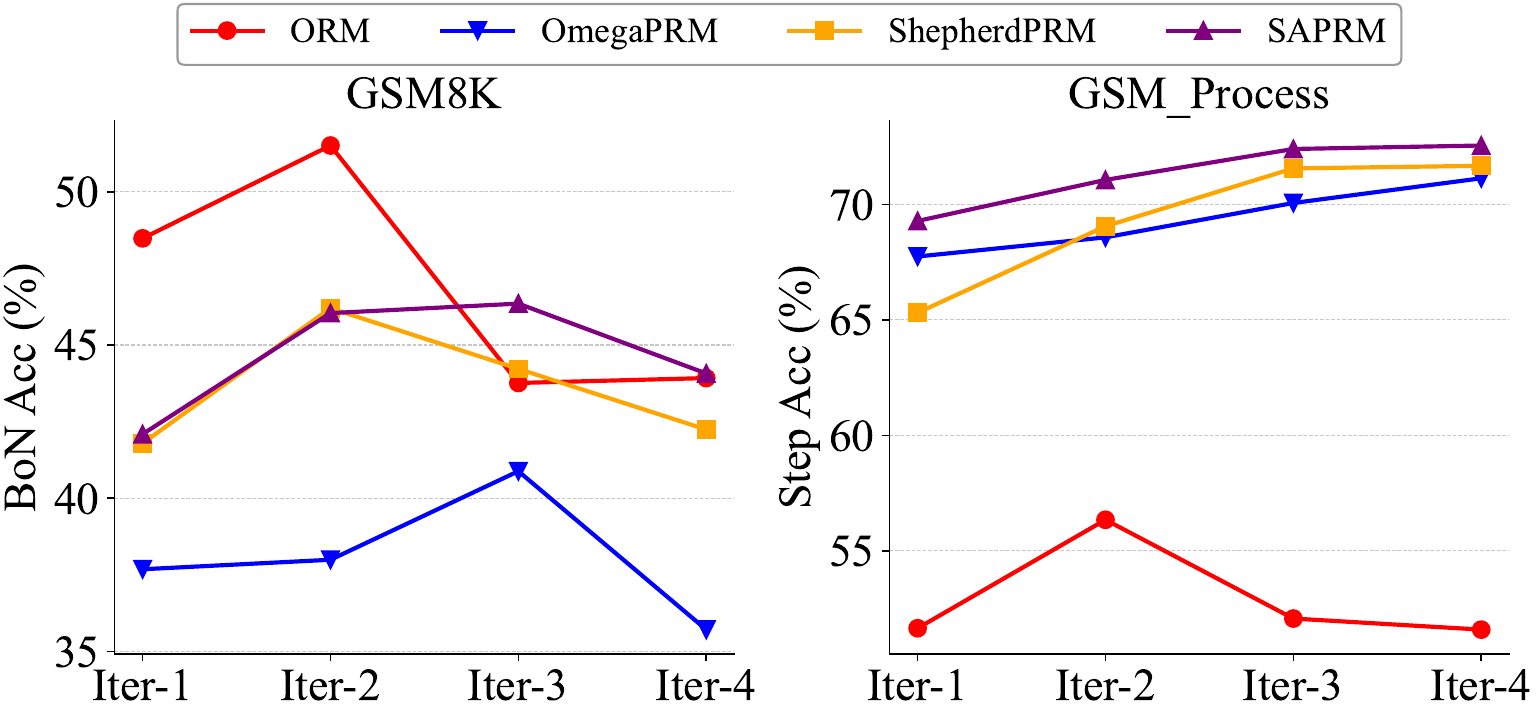}
  \caption{Performance Comparison of Different Verifier Models under BoN Acc and Step Acc. The base model selected for evaluation is Qwen-2.5-0.5B. For BoN evaluation \cite{cobbe2021training}, we utilize reasoning trajectories generated via self-sampling, ensuring that at least 50 distinct trajectories are generated per question.}
  \label{fig:fig7}
\end{figure*}

\begin{figure*}[!t]
  \centering
  \includegraphics[width=0.8\linewidth]{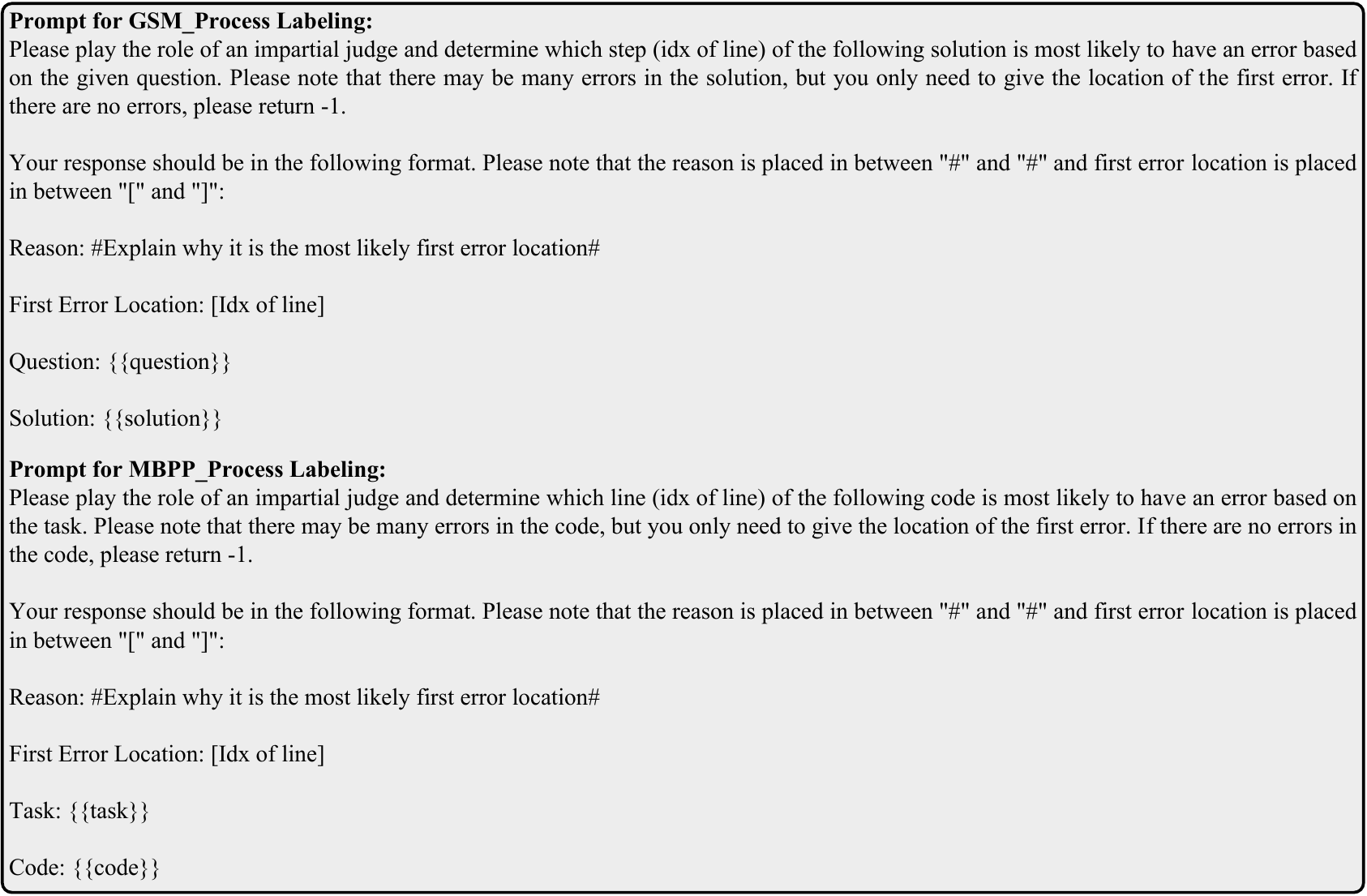}
  \caption{The prompts used for labeling GSM\_Process and MBPP\_Process both require GPT-4o to first provide a reason for identifying the first incorrect step, and then label the corresponding index of the step.}
  \label{fig:fig8}
\end{figure*}

To support this claim, we compare the performance of different reward models under two evaluation settings: final answer accuracy using the BoN approach and stepwise correctness prediction.

Figure \ref{fig:fig7} shows that all verification models experience performance degradation under BoN evaluation after multiple iterations. Although ORM initially demonstrates superior performance under BoN, it is eventually outperformed by other models in subsequent iterations.  In contrast, when evaluated based on the accuracy of step-wise label prediction, all types of PRM show a steady performance improvement. Therefore, BoN is not a reliable method for evaluating PRM.

\begin{figure*}[!t]
  \centering
  \includegraphics[width=0.8\linewidth]{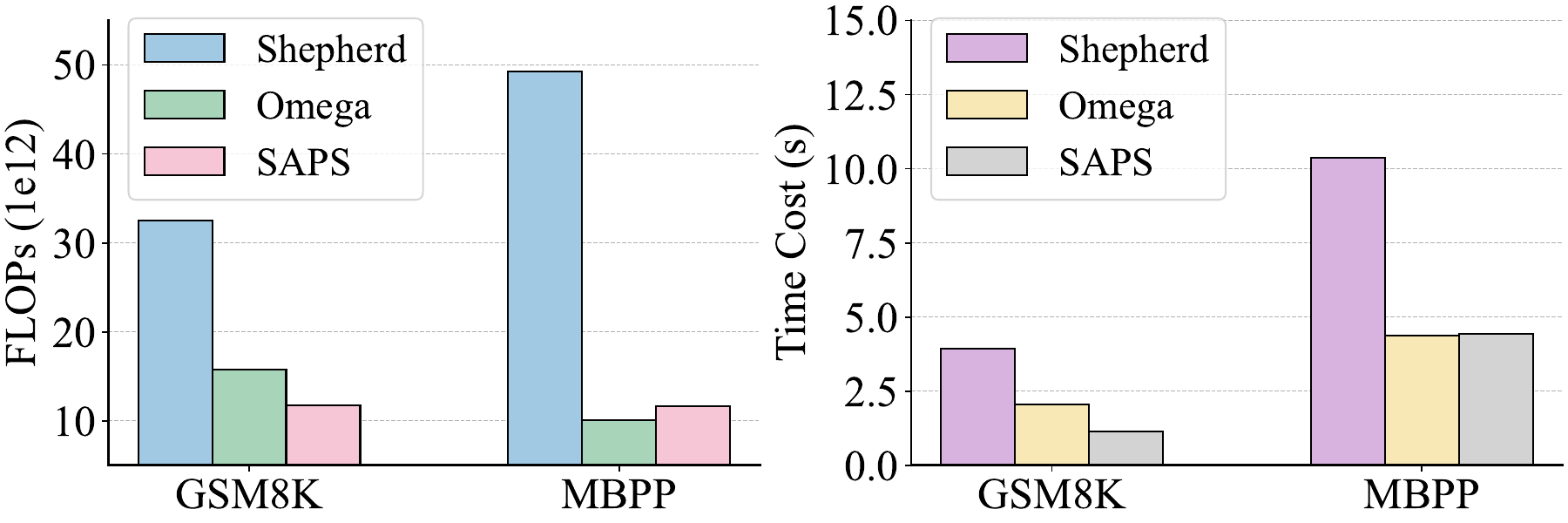}
  \caption{Comparison of efficiency in process labeling methods. The evaluation is based on sampled reasoning trajectories from Llama-3.2-1B, and the labeling is conducted on a single RTX 3090 GPU.}
  \label{fig:fig9}
\end{figure*}

Therefore, we propose GSM\_Process and MBPP\_Process, two step-level correctness prediction benchmarks designed respectively for mathematical reasoning and coding tasks. Although recent work has also introduced a similar benchmark, Process\_Bench \cite{zheng2024processbench}, we note that it is primarily designed for mathematical reasoning and is not well-suited for SLMs with limited domain knowledge.

Specifically, we collect reasoning trajectories from the fine-tuned Qwen-2.5-0.5B, Llama-3.2-1B, and Gemma-2-2B models on the GSM8K and MBPP test subsets, which contain 658 and 500 samples, respectively. Each model samples two trajectories per problem. Subsequently, GPT-4o is prompted to label each reasoning step based on the first error location annotation strategy. The prompt template used for the labeling process is illustrated in Figure \ref{fig:fig8}. The labeled samples are further filtered manually, primarily to remove the following cases: 1. Samples with correct final answers but containing incorrect steps; 2. Samples with incorrect final answers but no identifiable incorrect steps; 3. Samples where the reason for the labeling is flawed.

\section{C. Evaluation of Process Supervision Efficiency}
\label{apd:C}
The efficiency evaluation of different process supervision methods is conducted by measuring the average FLOPs and time (in seconds) required to annotate a single reasoning trajectory. The data used for evaluation is drawn from samples generated by the reasoning model itself. For GSM8K and MBPP, we randomly selected 400 and 200 problems, respectively, with each problem associated with 8 distinct reasoning trajectories.

\begin{figure*}[!t]
  \centering
  \includegraphics[width=0.8\linewidth]{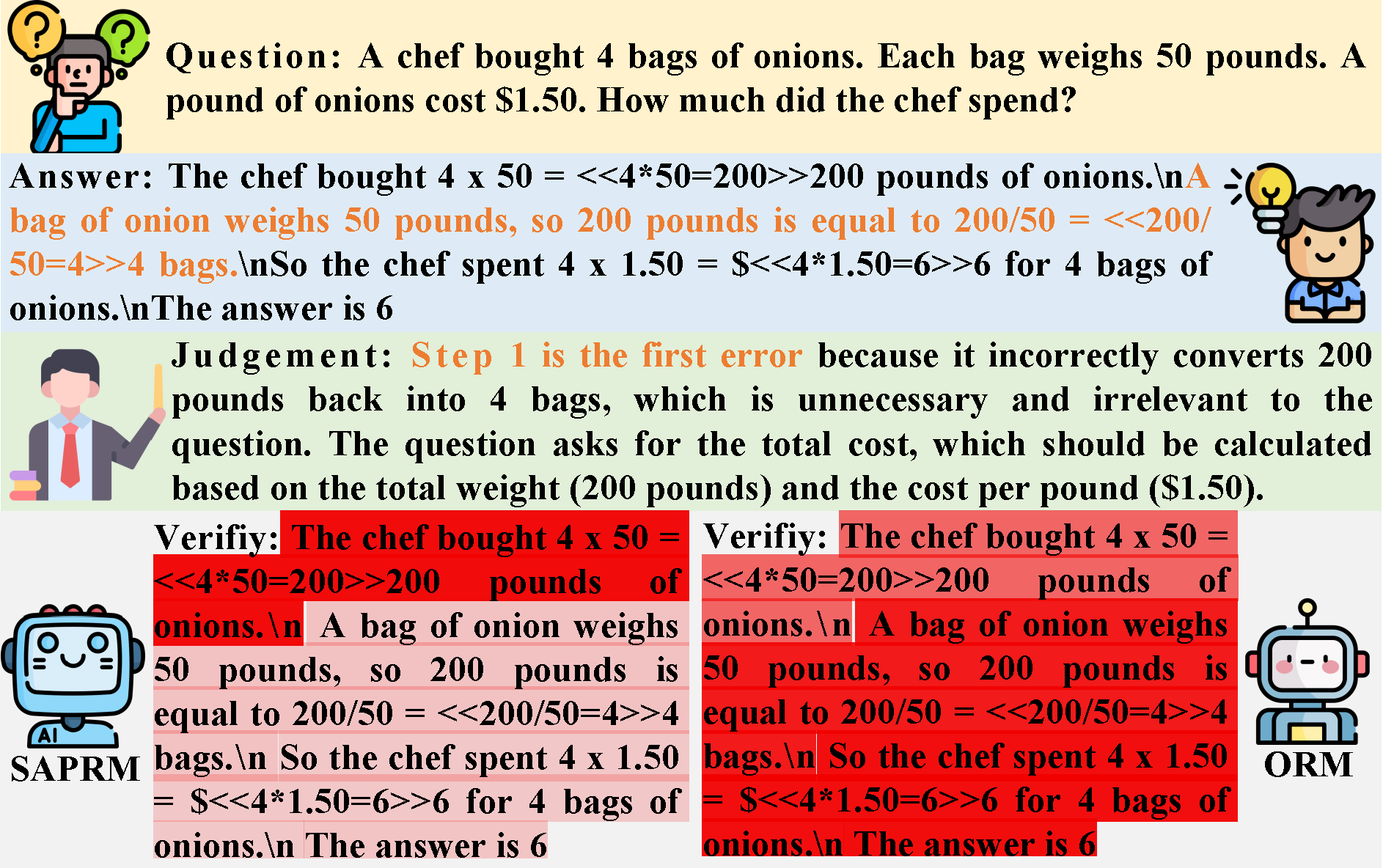}
  \caption{Case studies and comparisons of different verifiers (SAPRM vs. ORM) are presented. Judgements are provided by GPT-4o, indicating the correct first error position. In the visualization, a darker red color for a step indicates a higher score assigned by the verifier.}
  \label{fig:fig12}
\end{figure*}

Considering that evaluation results may vary across different hardware environments, all comparisons are conducted on a single NVIDIA 3090 GPU to ensure consistency. Additionally, in terms of time efficiency, the evaluation of MBPP includes the runtime of executing the test cases. Except for Figure \ref{fig:fig5}, Figure \ref{fig:fig9} also presents a comparison of efficiency among Shepherd \cite{DBLP:conf/acl/WangLSXDLCWS24}, Omega \cite{luo2024improve}, and SAPS when using Llama-3.2-1B as the reasoning models, respectively. The comparison of Llama further shows that SAPS is significantly more efficient than Shepherd, and the gap between them widens as the model size increases (e.g., Llama-3.2-1B vs. Qwen-2.5-0.5B). And SAPS performs comparably to or slightly better than Omega.

\section{D. Case Study of Process Verification}
\label{apd:D}
To further explore the specific performance of SAPRM, we conducted case studies comparing it with ORM, as shown in Figure \ref{fig:fig12}. The sampled reasoning paths are from GSM\_Process, with Llama-3.2-1B used as the base model. GPT-4o, serving as the judge, identified the correct first error in the reasoning trajectory at index 1. Based on this judgment, SAPRM assigned scores to different steps appropriately, giving a higher score to step 0 while assigning lower scores to the others. In contrast, ORM failed to distinguish between correct and incorrect steps, assigning high scores to all steps starting from step 1, despite the reasoning trajectory ultimately leading to an incorrect result.

\end{document}